\documentclass{article}
\pdfoutput=1

\usepackage[nonatbib,preprint]{neurips_2021}

\usepackage[utf8]{inputenc} 
\usepackage[T1]{fontenc}    
\usepackage{hyperref}       
\usepackage{url}            
\usepackage{booktabs}       
\usepackage{amsfonts}       
\usepackage{nicefrac}       
\usepackage{microtype}      
\usepackage{multirow}
\usepackage[labelfont=bf]{caption}
\usepackage{booktabs}
\usepackage{subcaption}
\usepackage{wrapfig}
\usepackage{xcolor}

\usepackage{amsmath}
\usepackage{amssymb}
\usepackage{MnSymbol}
\usepackage{algorithmic}
\usepackage[linesnumbered,ruled,noend]{algorithm2e}

\SetCommentSty{mycommfont}
\usepackage{cleveref}
\usepackage{enumitem}
\usepackage{color}
\usepackage{graphicx}
\usepackage{float}
\usepackage{bm}
\usepackage{bbm}
\usepackage{circledsteps}
\usepackage[normalem]{ulem}

\newcommand{\circled}[1]{\textcircled{\raisebox{-0.9pt}{#1}}}

\newcommand{\nrm}[1]{\textcolor[RGB]{0,0,0}{#1}}

\newcommand{\bst}[1]{\textcolor[RGB]{27,158,119}{\textbf{#1}}}

\newcommand{\phantomsubfigure}[1]{\begin{subfigure}[b]{0.0\textwidth}\phantomcaption\label{#1}\end{subfigure}}

\newcommand{\xhdr}[1]{\vspace{0.0mm}\noindent{\textbf{#1.}}\hspace{0.3mm}}
\newcommand{\xhdrr}[1]{\vspace{0.0mm}\noindent{\textbf{#1}}\hspace{0.3mm}}

\newcommand{\comment}[1]{}

\usepackage{hyperref}
\definecolor{mylinkcolor}{RGB}{0,0,170}
\hypersetup{colorlinks,allcolors=mylinkcolor,citecolor=mylinkcolor}

\usepackage{amsfonts}
\usepackage{amsmath}
\usepackage{mathrsfs} 
\usepackage{amssymb}
\usepackage{booktabs}
\usepackage{enumitem}
\usepackage{mathtools}

\newcommand{\mat}[1]{\bm{#1}}
\newcommand{\vct}[1]{\bm{#1}}
\newcommand{\var}{\mathrm{Var}}

\DeclareMathOperator{\argmin}{argmin}

\DeclareMathOperator{\E}{\mathbb{E}}

\newcommand{\mH}{\ensuremath{\mat{H}}}

\newcommand{\mX}{\ensuremath{\mat{X}}}

\newcommand{\vx}{\ensuremath{\vct{x}}}
\newcommand{\vy}{\ensuremath{\vct{y}}}

\newcommand{\cz}{\scriptsize}

\DeclarePairedDelimiterX{\dotp}[2]{\langle}{\rangle}{#1, #2}

\title{Graph Belief Propagation Networks}

\author{
  Junteng Jia \\
  Cornell University \\
  \href{mailto:jj585@cornell.edu}{jj585@cornell.edu} \\
    \And
  Cenk Baykal \\
  MIT \\
  \href{mailto:baykal@mit.edu}{baykal@mit.edu} \\
    \And
  Vamsi K. Potluru \\
  J.P. Morgan AI Research\\
  \href{mailto:vamsi.k.potluru@jpmchase.com}{vamsi.k.potluru@jpmchase.com} \\
    \And
  Austin R. Benson \\
  Cornell University \\
  \href{mailto:arb@cs.cornell.edu}{arb@cs.cornell.edu} \\
}

\begin{document}

\maketitle

%
\begin{abstract}
With the wide-spread availability of complex relational data, semi-supervised node classification in graphs has become a central machine learning problem.
Graph neural networks are a recent class of easy-to-train and accurate methods for this problem that map the features in the neighborhood of a node to its label, but they ignore label correlation during inference and their predictions are difficult to interpret.
On the other hand, collective classification is a traditional approach based on interpretable graphical models that explicitly model label correlations.
Here, we introduce a model that combines the advantages of these two approaches,
where we compute the marginal probabilities in a conditional random field, similar to collective classification, 
and the potentials in the random field are learned through end-to-end training, akin to graph neural networks.
In our model, potentials on each node only depend on that node's features, and edge potentials are learned via a coupling matrix.
This structure
enables simple training with interpretable parameters, 
scales to large networks,
naturally incorporates training labels at inference, 
and is often more accurate than related approaches.
Our approach can be viewed as either an interpretable message-passing graph neural network
or a collective classification method with higher capacity and modernized training.
\end{abstract}

\section{Bridging Graph Neural Networks and Collective Classification}
Graphs are a natural model for systems with interacting components,
where the nodes represent individual objects and the edges represent their interactions~\cite{easley2010networks,newman2010networks}.
For example, a social network might be modeled as a graph where the nodes are users and the edges are friendships.
Oftentimes, the nodes have attributes, such as a user's age, gender, or occupation in a social network,
%
but attributes are typically incomplete due to difficulties in data collection or privacy.
\textit{Graph-based semi-supervised learning}, also called \emph{node classification}, addresses this problem
by predicting a missing attribute (i.e., label $y$) on some nodes given other attributes (i.e., features $\vx$),
and has been used in a variety of applications, such as 
product category identification and protein function prediction~\cite{gallagher2008using,xu2010empirical,kipf2017semi,hamilton2017inductive,peel2017graph}.


Graph neural networks (GNNs) are a common method for semi-supervised learning on graphs~\cite{kipf2017semi,velickovic2018graph,wu2019simplifying,zhou2018graph}.
%
%
A GNN first summarizes the features and graph structure in the neighborhood of each node into a vector representation.
After, each node’s representation is used independently for classification.
Automatic differentiation enables end-to-end training, and there are simple sub-sampling schemes to handle large graphs~\cite{hamilton2017inductive}.
However, this approach implicitly assumes that node labels are conditionally independent given all features,
and standard GNNs do not (directly) use correlations between training and testing labels during inference.
%
%
Moreover, GNNs consist of transformation and aggregation functions parametrized by neural networks and the learned models are difficult to interpret.

On the other hand, collective classification (CC) is a class of interpretable methods based on graphical models that directly leverage label correlation for prediction~\cite{sen2008collective,jensen2004collective,london2014collective,neville2000iterative,lu2003link,macskassy2007classification,taskar2002discriminative}.
The statistical assumptions behind CC models are also arguably more appropriate than GNNs for graph data.
For instance, relation Markov networks (RMNs)~\cite{taskar2002discriminative} model the joint distribution of all node labels with a conditional random field and predict an unknown label with its marginal probabilities.
Leveraging label correlation during inference then simply amounts to conditioning on the training labels.
However, the interpretability and convenience come with a cost.
Collective classification models are learned by maximizing the joint likelihood, making end-to-end training extremely challenging.
The difficulty in learning model parameters, in turn, limits the capacity of the model itself.

The fact that GNNs and CC are two solutions to a same problem inspires a series of research questions.
 Is there a GNN architecture that learns the joint label distribution?
 Can we improve the modeling capacity and training of CC without hurting its interpretability?
 Finally, what type of approach would bridge those two largely independently developed fields?

Here, we answer these questions in the affirmative by developing graph belief propagation networks (GBPNs).
Our main idea is to learn the joint label distribution by maximizing the marginal likelihood of individual labels.
Specifically, we model the node labels in a graph as a conditional random field, 
where the coupling-potential on each edge is the same, 
but the self-potential on each node only depend on its features.
As such, our model is specified by a matrix for the coupling-potential that describes the affinity for nodes with certain labels to be connected, and a multi-layer perceptron (MLP) that maps a node's features to its self-potential.
To compute the marginal probabilities for each node label, we use loopy belief propagation (BP).
Putting everything together, we have a two-step inference algorithm that can be trained end-to-end: 
\circled{1} compute the self-potential on each node with an MLP, and 
\circled{2} estimate the marginal probabilities with a fixed number of BP iterations.

In one sense, GBPN is a message-passing GNN that projects node features to a dimension equal to the number of classes upfront, and subsequently use the projected features for propagation and aggregation.
On the other hand, GBPN learns the joint distribution of node labels in a graph similar to CC, and outputs the corresponding marginal probabilities for each node as a prediction.
%
%
The prediction accuracy is oftentimes higher than GNNs on benchmark datasets, and
the learned coupling matrix can be used to interpret the data and predictions. 
We also show this approach leads to straightforward mini-batch training so that the methods can be used on large graphs.

One issue is whether the marginal probabilities estimated by BP are good approximations on graphs with loops.
Empirically, BP often converges on graphs with cycles with estimated marginals that are close to the ground truth~\cite{murphy1999loopy}.
In our context, we find that BP converges in just a few iterations on several real-world graph datasets.
%
Another possible concern is whether learning a joint distribution by maximizing the marginal likelihood produces good estimates.
We show that on synthetic sampled from Markov random fields (MRFs), 
this approach recovers the parameters used to generate the data.

%
%


\subsection{Additional related work}

\xhdr{Collective classification}
Collective classification encompasses many 
machine learning algorithms for relational data that use correlation in
labels among connected entities in addition to features on the entities~\cite{sen2008collective,jensen2004collective,london2014collective}.
For example, local conditional classifiers iteratively make predictions and update node features based on these predictions~\cite{neville2000iterative,lu2003link,macskassy2007classification,mcdowell2007cautious}.
%
Closer to our methods are approaches that use pairwise MRFs
and belief propagation~\cite{taskar2002discriminative}.
The major difference is that they estimate the MRF coupling-potentials by
maximizing the joint likelihood of all the nodes, and therefore require a fully
labeled graph for training (see \Cref{sec:comparisons}).
In contrast, our model is learned discriminatively end-to-end on the training
nodes by back-propagating through belief propagation steps.

\xhdr{MRFs with GNNs}
A few approaches combine MRFs and graph neural
networks~\cite{qu2019gmnn,ma2019cgnf,gao2019conditional}.
They add a conditional random field layer to model the joint distribution of the
node labels and maximize the joint probability of the observed labels
during training.
Variational methods or pseudolikelihood approaches are used to optimize the joint likelihood.
Our GBPN model avoid these difficulties by optimizing for the marginal likelihood,
and this leads to more accurate predictions.

\xhdr{Direct use of training labels}
Besides MRFs, other GNN approaches have used training
labels at inference in graph-based learning. These include diffusions on
training labels for data augmentation~\cite{shi2020masked} and post-processing
techniques~\cite{jia2020residual,huang2021combining,jia2021unifying}. Similar in spirit,
smoothing techniques model positive label correlations in
neighboring nodes~\cite{klicpera2018predict,bojchevski2020scaling}. Our
approach is more statistical, as GBPN just conditions on known
labels. 

\xhdr{Learning better belief propagation algorithms}
Some recent research learns a belief propagation algorithm to estimate the
partition function in a graphical
model~\cite{yoon2018inference,kuck2020belief}. 
In comparison, we focus on the node classification problem for real-world graphs.


\section{Node Classification as Statistical Inference in a Markov Random Field}
%
%
Consider an attributed graph $G(V, E, \mX, \vy)$ where $V$ is a set $n$ of nodes, $E$ is a set of undirected edges, $\mX \in \mathbb{R}^{n \times d}$ is a matrix whose rows correspond to node features, and $\vy \in \{0, \ldots, c-1\}^{n}$ is a vector of node labels in one of $c$ possible classes.
We partition the vertices into two disjoint sets $L \cup U = V$, where $L$ is the training set (labeled) and $U$ is the testing set (unlabeled).
In the \emph{node classification} task, we assume access to all node features $\{\vx_{i}\}_{i \in V}$ and the labels on the training nodes $\{y_{i}\}_{i \in L}$, and the goal is to predict the testing labels $\{y_{i}\}_{i \in U}$.
The basic idea of our approach is to define a joint distribution over node features and labels $p(\mX, \vy)$ and
then cast the node classification task as computing the marginal probability $p(y_{i} | \mX, \vy_{L})$.

\subsection{A Data Generation Process for Node Attributes \label{subsec:generative_model}}
We treat the graph as deterministic but the node attributes randomly as follows:
\circled{1} Jointly sample the node labels $p(\vy)$;
\circled{2} Given the node labels $\vy$, sample the features from $p(\mX|\vy)$.
This generation process defines a joint distribution $p(\mX, \vy)$.
Furthermore, we add the following assumptions.
\begin{itemize}[leftmargin=0.20in]
    \vspace{-0.07in}
    \setlength\itemsep{0.03in}
    \item First, the joint distribution of node labels $p(\vy)$ is a pairwise Markov random field,
    \begin{align}
\textstyle        p(\vy) = \frac{\varphi(\vy)}{\sum_{\vy'} \varphi(\vy')}, \quad \varphi(\vy) = \prod_{i \in V} h_{i}(y_{i}) \prod_{(i,j) \in E} H_{ij}(y_{i}, y_{j}),
    \end{align}
    and the coupling-potential on every edge is the same and symmetric, i.e. $H_{ij}(y_{i}, y_{j}) = H_{y_{i}, y_{j}} = H_{y_{i}, y_{j}}$, where $\mH \in \mathbb{R}_{+}^{c \times c}$.
    This assumption implies each node is directly coupled only with its neighbors, a condition known as the Markov propriety in probabilistic graph models~\cite{koller2009probabilistic}.
    \item Second, The conditional distribution $p(\mX|\vy)$ can be factorized as
    \begin{align}
    \textstyle        p(\mX|\vy) = \prod_{i \in V} p(\vx_{i}|y_{i}) = \prod_{i \in V} f(y_{i}; \vx_{i}),
    \end{align}
    which means the features on each node is independently sampled given its label.
    \vspace{-0.05in}
\end{itemize}
These two assumptions lead to simple and efficient inference algorithms, as we will show next.

\subsection{Node Classification Models Derived through Probabilistic Inference \label{subsec:inference_algorithm}}
Following the generative model defined above,  we derive the posterior probability as:
\begin{align}
    \textstyle    p(\vy|\mX) = \frac{p(\mX, \vy)}{p(\mX)} = \frac{p(\vy) p(\mX|\vy)}{\sum_{\vy'} p(\vy') p(\mX|\vy')} &\cong \textstyle \prod_{i \in V} f(y_{i}; \vx_{i})\ h_{i}(y_{i}) \prod_{(i,j) \in E} H_{ij}(y_{i}, y_{j}) \\
    &= \textstyle \prod_{i \in V} g(y_{i}; \vx_{i}) \prod_{(i,j) \in E} H_{ij}(y_{i}, y_{j}),
    \label{eq:posterior_distribution}
 \end{align}
which is a conditional random field~\cite{lafferty2001conditional}, where $\cong$ denotes ``equality up to a normalization''~\cite{mezard2009information}.
We directly parametrize this posterior distribution using a symmetric matrix $\mH$ for the coupling-potential, and a neural network function $g_{\theta}$ to map from a node's features to its self-potential.
Similar modeling assumptions have been considered in collective classification~\cite{taskar2002discriminative,xiang2008pseudolikelihood}, where the models are trained to maximize the joint likelihood.
In contrast, our model is trained on the marginal likelihood of individual labels.
Next, we introduce the inference algorithm for the marginal probability $p(y_{i}|\mX)$.
The algorithm only use node features for label prediction, and (once trained) can be used to predict labels in a unseen graph.
Later, we consider the inference algorithm for $p(y_{i}|\mX,\vy_{L})$, which gives more accurate predictions in the transductive learning setting.

\xhdr{Inference}
Exact inference for the marginal probabilities is intractable on large-scale graphs with cycles~\cite{koller2009probabilistic}.
Therefore, we resort to approximate inference using the loopy belief propagation (BP) algorithm~\cite{murphy1999loopy}.
%
In particular, we estimate the marginal probabilities using a two-step algorithm.
\begin{enumerate}[leftmargin=0.2in]
    \item[\circled{1}] Compute the self-potentials using a multi-layer perceptron (MLP) with parameters $\theta$,
    \begin{align}
        g_{\theta}(y_{i}; \vx_{i}) = \texttt{Cat}(y_{i}|\texttt{softmax}(\texttt{MLP}_{\theta}(\vx_{i}))).
    \end{align}
    \item[\circled{2}] Run $T$ iterations of belief-propagation,
    \begin{align}
\textstyle         p_{i}^{(t)}(y_{i}) \cong p_{i}^{(0)}(y_{i}) \prod_{j \in N(i)} m_{ji}^{(t)}(y_{i}), \quad m_{ji}^{(t)}(y_{i}) \cong \sum_{y_{j}} H_{ji}(y_{j}, y_{i}) \frac{p_{j}^{(t-1)}(y_{j})}{m_{ij}^{(t-1)}(y_{j})},
    \end{align}
    where $t = 1, \ldots, T$ and $N(i)$ are the neighbors of node $i$.
    The initial condition is given by,
    \begin{align}
        p_{i}^{(0)}(y_{i}) = g(y_{i}; \vx_{i}), \quad m_{ji}^{(0)}(y_{j}, y_{i}) = \nicefrac{1}{c}.
    \end{align}
\end{enumerate}
In other words, we set the initial belief on each node to its self-potential and update it with messages from the neighbors.
The final belief $p_{i}(y_{i}) = p_{i}^{(T)}(y_{i})$ is used to approximate $p(y_{i}|\mX)$.

In practice, for numerical stability, we perform belief-propagation updates in the log-space.
\begin{align}
\textstyle    \log p_{i}^{(t)}(y_{i}) &\cong \log p_{i}^{(0)}(y_{i}) + \sum_{j \in N(i)} \log m_{ji}^{(t)}(y_{i}) \label{eq:gbpn_aggregation}\, , \\
\textstyle    \log m_{ji}^{(t)}(y_{i}) &\cong \texttt{LSE}_{y_{j}}\left[\log H_{ji}(y_{j}, y_{i}) + \log p_{j}^{(t-1)}(y_{j}) - \log m_{ij}^{(t-1)}(y_{j})\right],
    \label{eq:gbpn_message}
\end{align}
where $\texttt{LSE}$ stands for the log-sum-exp function: $\texttt{LSE}_{y_{j}}\left[f(y_{i},y_{j})\right] = \log\left[\sum_{y_{j}} \exp\left(f(y_{i}, y_{j})\right)\right].$

\xhdr{Learning}
We optimize the model by maximizing the log marginal-likelihood of the training labels,
\begin{align}
\textstyle    \theta^{*}, \mH^{*} = \arg\max_{\theta, \mH} \sum_{i \in L} \log p(y_{i}|\mX) \approx \arg\max_{\theta, \mH} \sum_{i \in L} \log p_{i}(y_{i}).
    \label{eq:gbpn_loss}
\end{align}
The gradients with respective to the parameters are computed with auto-differentiation by backpropagating through the BP iterations.
Since the inference algorithm only uses node features, we can learn the parameters in one graph with abundant labels, and predict for another graph where the labels are difficult to obtain.
As such, we denote this algorithm as the inductive variant of GBPN, or GBPN-I.

\xhdr{Direct use of training labels for inference}
In the transductive learning setting, both the training and testing nodes are from the same graph.
Therefore, we can directly leverage the training labels during inference.
From the probabilistic inference perspective, we start with a joint distribution over testing labels by conditioning on both the features and the training labels, i.e.,
\begin{align}
    p(\vy_{U}|\mX,\vy_{L}) \cong \prod_{i \in U} \left[g(y_{i}; \vx_{i}) \prod_{i \in U,\ j \in L,\ (i,j) \in E} H_{ij}(y_{i}, y_{j})\right] \prod_{i, j \in U,\ (i,j) \in E} H_{ij}(y_{i}, y_{j}).
    \label{eq:posterior_distribution_with_training_labels}
\end{align}
This is nothing but another conditional random field defined on the induced subgraph $G[U]$, where the self-potentials are modified to incorporate the couplings between the training and testing nodes in the original graph.
Then, similar to the inductive algorithm, we estimate the marginal probabilities by belief propagation.
To better mimic how we predict the testing labels, during each training step, we randomly select half of the training nodes to be conditioned on, and optimize the log-likelihood loss on the other half.
We denote this transductive variant of inference and training as GBPN.

\subsection{Comparison with Other Node Classification Methods \label{sec:comparisons}}
Now that our method is fully specified, we comparing it with existing methods in terms of statistical assumption, model parametrization, training/inference objective and optimization algorithm.

\xhdrr{GNNs}~\cite{zhou2020graph} independently predict the label on each node using node features in the neighborhood.
The implicit statistical assumption is that node labels are conditional independent given the features:
\begin{align}
    \textstyle p(\vy|\mX) = \prod_{i \in V} p(y_{i}|\mX)
    \label{eq:gnn_posterior_distribution}
\end{align}
Therefore, GNNs ignore label correlation and directly parametrize $p(y_{i}|\mX)$ to extract the most information from features.
In contrast, GBPN models $p(\vy|\mX)$ and uses label correlation for inference.

\xhdrr{RMN}~\cite{taskar2002discriminative} is a representative example for collective classification methods~\cite{sen2008collective} that directly maximize the joint likelihood $p(\vy|\mX)$.
Those models only work for the inductive setting, where they are first trained on a fully labeled graph and then used to predict node labels in a different graph.
In comparison, our model can be trained on a partially labeled graph, and it is suitable for both the inductive and transductive setting.
Moreover, even on a graph that is fully labeled, methods like RMN are historically difficult to train.
This is because the gradient of the joint likelihood involves a normalization factor and is therefore hard to compute.
GBPN avoids this problem by differentiating through the belief-propagation iterations with backpropagation.

\xhdrr{Pseudolikelihood EM}~\cite{xiang2008pseudolikelihood}
was designed to learn the correlation structure in a partially labeled graph
by maximizing the evidence lower bound (ELBO) for the joint likelihood of the training labels $p(\vy_{L}|\mX)$. 
Maximizing the ELBO boils down to directly optimizing two marginal distributions: a variational distribution $q_{\theta}$ and a pseudolikelihood $p_{\phi}$ used to approximate the posterior.
Therefore, if $p_{\phi}$ and $q_{\theta}$ are neural networks (as in the GMNN model~\cite{qu2019gmnn}), then the entire model can be trained with auto-differentiation.
The learned $p_{\phi}$ can leverage training labels during inference, akin to GBPN.
However, while GMNN models label correlation with neural networks, GBPN explicitly specifies the coupling-potential and is thus more interpretable.
Moreover, we compare GBPM with GMNN in \cref{subsec:results_full_batch} and find GBPN gives higher prediction accuracies on benchmark datasets.
\begin{table}[t]
    \caption{Summary of node classification methods from the probabilistic inference perspective.
    }
    \centering
    \resizebox{1.0\linewidth}{!}{
    \begin{tabular}{r @{\quad} lllll}
    \toprule
        method    &  assumption                             &  model parametrization                                    &  training                  &  inference                &  optimization        \\
    \midrule                                                                                                                                                                     
        GBPN      &  \cref{eq:posterior_distribution}       &  $p(\vy|\mX)$                                             &  $p(y_{i}|\mX, \vy_{L})$   &  $p(y_{i}|\mX, \vy_{L})$  &  auto-diff           \\
        GNN       &  \cref{eq:gnn_posterior_distribution}   &  $p(y_{i}|\mX)$                                           &  $p(y_{i}|\mX)$            &  $p(y_{i}|\mX)$           &  auto-diff           \\
        RMN       &  \cref{eq:posterior_distribution}       &  $p(\vy|\mX)$                                             &  $p(\vy|\mX)$              &  $p(y_{i}|\mX, \vy_{L})$  &  manual-diff         \\
        GMNN      &  \cref{eq:posterior_distribution}       &  $p_{\phi}(y_{i}|\mX, \vy_{N(i)}), q_{\theta}(y_{i}|\mX)$ &  $p(\vy_{L}|\mX)$          &  $p(y_{i}|\mX, \vy_{L})$  &  EM (auto-diff)      \\
    \bottomrule
    \end{tabular}
    }
    \label{tab:comparisons}
\end{table}

\subsection{Practical Considerations for Training GBPNs \label{subsec:practical_considerations}}
\begin{wrapfigure}{r}{0.33\textwidth}
    \vspace{-0.18in}
    \centering
    \includegraphics[width=0.9\linewidth]{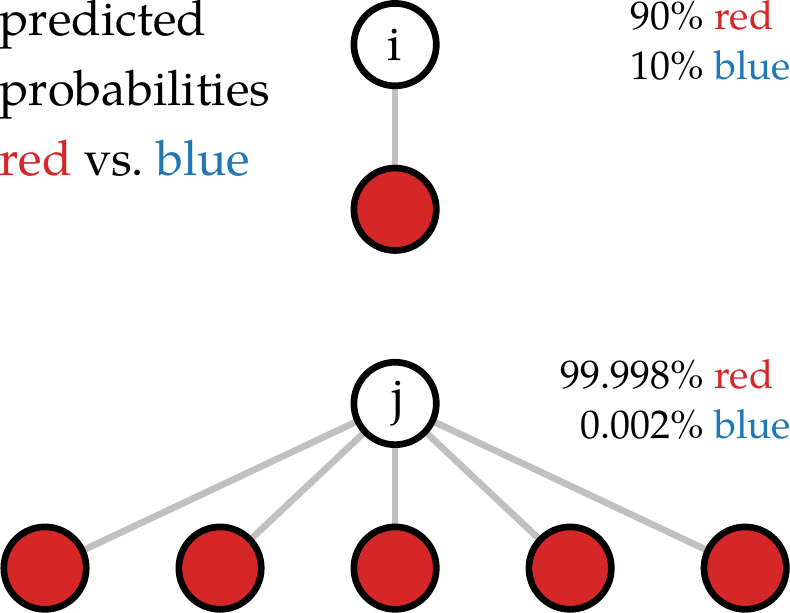}
    \caption{Predicted probabilities for low- \textit{vs.} high-degree nodes.}
    \label{fig:heterogeneity}
    \vspace{-0.18in}
\end{wrapfigure}
One modeling choice we have to make for GBPN is the loss function for training.
%
Although the (unweighted) negative log-likelihood loss is often used for training GNNs,
the same choice for GBPN can lead to sub-optimal performance in practice, 
especially when the degree heterogeneity in a graph is large.
The reason is that, unlike GNNs, the predictive confidence of GBPN on a node can grow exponentially with the node degree, as shown in \cref{fig:heterogeneity}.
Therefore, the negative log-likelihood loss associated with mis-classifying a node grows linearly with the node degree.
This creates a potential misalignment between the loss function and the classification accuracy, 
as classification accuracy is not weighted based on node degree.
We will demonstrate this misalignment empirically in \Cref{subsec:subsampling}.

We employ a straightforward way to align the loss function to the accuracy measure, which is to reweight the marginal likelihood in the loss function \cref{eq:gbpn_loss} as $\log p_{i}'(y) \cong \alpha_{i} \cdot \log p_{i}(y)$.
Here, $\alpha_{i}$ reduces the influence of high-degree nodes.
In practice, we find that $\alpha_{i} = d(i)^{-1/2}$ gives consistently better performance than $\alpha_{i} = 1$ (no reweighting), where $d(i)$ is the degree of node $i$.

One limitation of GBPN is that it requires storing all the messages between neighbors during training, which costs $\mathcal{O}(|E|)$ per BP step.
In comparison, GCN~\cite{kipf2017semi} and GraphSAGE~\cite{hamilton2017inductive} costs $\mathcal{O}(|V|)$ per layer, and GAT~\cite{velickovic2018graph} costs $\mathcal{O}(|E|)$ per layer due to edge attention weights.
To reduce the computational cost on large-scale network, we will introduce a mini-batch training algorithm for GBPN in \Cref{subsec:subsampling}.


\section{Experiments \label{sec:experiments}}
Now, we compare GBPN against GNNs empirically on several synthetic and real-world datasets,
and also interpret the learned GBPN models, as well as discuss new sub-sampling techniques for large networks.
The proposed models are summarized with pseudocode in \cref{subsec:additional_implementation} and implemented in PyTorch Geometric~\cite{fey2019fast}.
The source code, data, and experiments are publicly available online at \url{https://github.com/000Justin000/GBPN.git}.
Statistics of all the datasets are in \cref{tab:data_statistics} and we describe them below.

%
\begin{wraptable}{r}{0.53\textwidth}
\vspace{-12mm}
\begin{minipage}{1.00\linewidth}
\begin{table}[H]
    \caption{Summary of dataset statistics.}
    \centering
    \resizebox{1.00\textwidth}{!}{
    \begin{tabular}{r @{\quad} rrrr}
    \toprule
              &  \# nodes                & \# edges                & \# features  & \# classes \\
    \midrule
    Ising$+$  &  2.6\cz{$\times 10^3$}  & 1.0\cz{$\times 10^4$}  & 2           & 2        \\
    Ising$-$  &  2.6\cz{$\times 10^3$}  & 1.0\cz{$\times 10^4$}  & 2           & 2        \\
      MRF$+$  &  2.6\cz{$\times 10^3$}  & 1.0\cz{$\times 10^4$}  & 2           & 3        \\
      MRF$-$  &  2.6\cz{$\times 10^3$}  & 1.0\cz{$\times 10^4$}  & 2           & 3        \\
    \midrule
        Cora  &  2.7\cz{$\times 10^3$}  & 5.3\cz{$\times 10^3$}  & 1433        & 7        \\ 
    CiteSeer  &  3.3\cz{$\times 10^3$}  & 4.6\cz{$\times 10^3$}  & 3703        & 6        \\ 
      PubMed  &  2.0\cz{$\times 10^4$}  & 4.4\cz{$\times 10^4$}  & 500         & 3        \\ 
          CS  &  1.8\cz{$\times 10^4$}  & 8.2\cz{$\times 10^4$}  & 6805        & 15       \\ 
     Physics  &  3.4\cz{$\times 10^4$}  & 2.5\cz{$\times 10^5$}  & 8415        & 5        \\ 
    Election  &  3.1\cz{$\times 10^3$}  & 2.3\cz{$\times 10^3$}  & 9           & 2        \\ 
      Sexual  &  1.9\cz{$\times 10^3$}  & 2.1\cz{$\times 10^3$}  & 20          & 2        \\ 
    \midrule
    Elliptic  &  2.0\cz{$\times 10^5$}  & 2.3\cz{$\times 10^5$}  & 165         & 2        \\ 
    Payments  &  2.2\cz{$\times 10^5$}  & 2.8\cz{$\times 10^5$}  & 573         & 2        \\ 
       arXiv  &  1.7\cz{$\times 10^5$}  & 1.2\cz{$\times 10^6$}  & 128         & 40       \\ 
    Products  &  2.4\cz{$\times 10^6$}  & 6.2\cz{$\times 10^7$}  & 100         & 47       \\ 
    \bottomrule
    \end{tabular}
    }
    \label{tab:data_statistics}
\end{table}
\end{minipage}
\vspace{-7mm}
\end{wraptable} 

\xhdr{Synthetic MRF datasets}
We sample data from Markov random fields, i.e., from our generative model.
Specifically, we use a $51 \times 51$ grid graph, where each node belongs to one of two possible classes.
The neighboring node labels are likely to be the same if the coupling-potential is positive (denoted $+$), and likely to be different if the coupling-potential is negative (denoted $-$).
This setup is known as the Ising model in statistical physics and models
homophily or heterophily in social networks~\cite{mcpherson2001birds},
or assortativitity and disassortativity more generally~\cite{newman2002assortative}.
%
%
We also consider a setup with three classes with either positive or negative coupling-potentials.
In all settings, we use the grid coordinates as features to predict labels.

\xhdr{Citation graphs}
We use the small benchmark graphs Cora, CiteSeer and PubMed~\cite{sen2008collective}, 
as well as the larger arXiv benchmark~\cite{hu2020ogb},
where the vertices represent articles and edges represent citations. 
Features are derived from the article text, and the goal is to predict the research field of each article.

\xhdr{Co-authorship graphs}
These are graphs of computer scientists and physicists~\cite{shchur2018pitfalls}, 
where nodes are researchers, 
edges connect researchers that have coauthored a paper,
node features are paper keywords for each author, and
the goal is to predict the most active field of study for each author.

\xhdr{U.S. election}
This dataset comes from the 2016 presidential election, where nodes are U.S. counties, and edges connect counties with the strongest Facebook social ties~\cite{jia2021unifying}.
Each node has county-level demographic features (e.g., median income) and social network user features (e.g., fraction of friends within 50 miles).
The goal is to predict the party (Republican or the Democrat) that won each county.

\xhdr{Sexual interactions}
This is a social network where connections come from sexual interactions~\cite{jia2020residual,morris2011hiv}.
Node features include occupation and retirement status, and the goal is to predict gender.

\xhdr{Financial transactions}
We use two financial networks, where the nodes represent transactions and the edges are payment flows.
Node features are metadata associated with each transaction.
The Elliptic dataset is derived from bitcoin transactions~\cite{pareja2020evolvegcn},
and transactions are labeled as licit or illicit.
The Payments dataset consists of synthetic payments that resemble real-world data,
provided by J.P.~Morgan Chase \& Co.
Transactions are labeled as fraudulent or non-fraudulent.

\xhdr{Co-purchasing}
%
%
Here, nodes are products on Amazon, and edges connect co-purchased products.
The features are the average word embeddings from the descriptions and labels are product categories.

\subsection{Performance on Smaller Datasets \label{subsec:results_full_batch}}
The datasets we consider here fits into three groups: \circled{1} small synthetic graphs; \circled{2} small real-world graphs; \circled{3} large-scale graphs.
We first consider \circled{1} and \circled{2}, where training the entire graph can fit into a single GPU.
On these datasets, all models take at most 20 minutes to train on a Telsa V100 GPU.

%
For GBPN, we use a $2$-layer MLP with $256$ hidden units to map the features of each node to its self-potentials.
Then we run five steps of BP iterations to compute the predictions.
For standard baselines, we consider a $2$-layer MLP, GCN, and GraphSAGE models with the $256$ hidden units, 
as well as a $2$-layer GAT with four attention heads $\times$ $64$ hidden units per layer.
We also compare GBPN with GMNN~\cite{qu2019gmnn}, 
which is also based on Markov random fields, as well as
DeeperGNN~\cite{liu2020towards}, which uses several hops of neighborhood information at anode.
For both of these baselines, we use the reference implementations\footnote{DeeperGNN is released under GNU General Public License while GMNN is under no license.} with default hyperparameters for training and inference.

We randomly split all datasets into 30\% training, 20\% validation, and 50\% testing.
All baseline methods are trained by minimizing the negative log-likelihood (NLL), while GBPN is trained with the weighted NLL loss discussed in \Cref{subsec:practical_considerations}.
For all methods, we use an AdamW optimizer with learning rate $1.0 \times 10^{-3}$ and decay rate $2.5 \times 10^{-4}$ to perform full-batch training for 500 steps.
The testing accuracies are averaged over 30 runs and summarized in \Cref{tab:results_synthetic_full_batch,tab:results_empirical_full_batch}.
\begin{table}[t]
    \caption{Node classification performances on synthetic data with full-batch training. The synthetic is sampled from our generative model, and, our algorithms are appropriately more accurate.}
    \centering
    \resizebox{1.0\linewidth}{!}{
    \begin{tabular}{r @{\qquad} |llllll|ll}
    \toprule
        dataset       &  MLP                        &  GCN                        &  SAGE                       &  GAT                        &  GMNN                       &  DeeperGNN                  &  GBPN-I                     &  GBPN                        \\
    \midrule                                                                                                                                                                
        Ising$+$      &  \nrm{67.1} {\cz $\pm$ 1.6} &  \nrm{67.1} {\cz $\pm$ 1.8} &  \nrm{68.2} {\cz $\pm$ 1.5} &  \nrm{64.4} {\cz $\pm$ 2.2} &  \nrm{64.4} {\cz $\pm$ 3.1} &  \nrm{65.6} {\cz $\pm$ 1.6} &  \nrm{68.5} {\cz $\pm$ 1.8} &  \bst{75.0} {\cz $\pm$ 1.4}  \\
        Ising$-$      &  \nrm{48.8} {\cz $\pm$ 0.8} &  \nrm{48.7} {\cz $\pm$ 0.7} &  \nrm{48.4} {\cz $\pm$ 1.0} &  \nrm{49.3} {\cz $\pm$ 0.9} &  \nrm{49.4} {\cz $\pm$ 0.9} &  \nrm{49.3} {\cz $\pm$ 1.1} &  \nrm{48.3} {\cz $\pm$ 1.1} &  \bst{72.7} {\cz $\pm$ 1.4}  \\
          MRF$+$      &  \nrm{64.2} {\cz $\pm$ 1.9} &  \nrm{65.9} {\cz $\pm$ 2.0} &  \nrm{64.8} {\cz $\pm$ 1.9} &  \nrm{61.4} {\cz $\pm$ 2.3} &  \nrm{45.8} {\cz $\pm$ 2.9} &  \nrm{53.3} {\cz $\pm$ 3.8} &  \nrm{66.5} {\cz $\pm$ 2.0} &  \bst{70.3} {\cz $\pm$ 1.7}  \\
          MRF$-$      &  \nrm{64.3} {\cz $\pm$ 4.1} &  \nrm{66.3} {\cz $\pm$ 5.0} &  \nrm{65.3} {\cz $\pm$ 4.6} &  \nrm{62.3} {\cz $\pm$ 4.0} &  \nrm{41.3} {\cz $\pm$ 5.3} &  \nrm{53.1} {\cz $\pm$ 2.9} &  \nrm{66.3} {\cz $\pm$ 5.0} &  \bst{73.5} {\cz $\pm$ 4.5}  \\
    \bottomrule
    \end{tabular}
     }
    \label{tab:results_synthetic_full_batch}
\end{table}
\begin{table}[t]
    \caption{Node classification performances on smaller empirical datasets with full-batch training.}
    \centering
    \resizebox{1.0\linewidth}{!}{
    \begin{tabular}{r @{\qquad} |llllll|ll}
    \toprule
        dataset       &  MLP                        &  GCN                        &  SAGE                       &  GAT                        &  GMNN                       &  DeeperGNN                  &  GBPN-I                     &  GBPN                       \\
    \midrule                                                                                                                                                                
        Cora          &  \nrm{72.1} {\cz $\pm$ 1.3} &  \nrm{87.1} {\cz $\pm$ 0.7} &  \nrm{86.9} {\cz $\pm$ 0.8} &  \nrm{87.1} {\cz $\pm$ 0.9} &  \nrm{86.4} {\cz $\pm$ 0.9} &  \bst{87.2} {\cz $\pm$ 0.8} & \nrm{85.6} {\cz $\pm$ 0.7} &  \nrm{86.4} {\cz $\pm$ 0.9}  \\
        CiteSeer      &  \nrm{71.2} {\cz $\pm$ 0.9} &  \nrm{73.5} {\cz $\pm$ 1.0} &  \nrm{73.2} {\cz $\pm$ 1.0} &  \nrm{73.1} {\cz $\pm$ 1.2} &  \nrm{72.9} {\cz $\pm$ 1.2} &  \nrm{73.9} {\cz $\pm$ 0.8} & \nrm{74.7} {\cz $\pm$ 1.3} &  \bst{74.8} {\cz $\pm$ 0.8}  \\
        PubMed        &  \nrm{86.5} {\cz $\pm$ 0.2} &  \nrm{87.1} {\cz $\pm$ 0.3} &  \nrm{87.8} {\cz $\pm$ 0.4} &  \nrm{88.1} {\cz $\pm$ 0.3} &  \nrm{86.7} {\cz $\pm$ 0.2} &  \nrm{84.7} {\cz $\pm$ 0.3} & \nrm{88.4} {\cz $\pm$ 0.3} &  \bst{88.5} {\cz $\pm$ 0.3}  \\
        CS            &  \nrm{94.2} {\cz $\pm$ 0.2} &  \nrm{93.2} {\cz $\pm$ 0.2} &  \nrm{93.7} {\cz $\pm$ 0.2} &  \nrm{94.0} {\cz $\pm$ 0.3} &  \nrm{93.3} {\cz $\pm$ 0.3} &  \nrm{94.9} {\cz $\pm$ 0.2} & \bst{95.5} {\cz $\pm$ 0.2} &  \bst{95.5} {\cz $\pm$ 0.3}  \\
        Physics       &  \nrm{95.8} {\cz $\pm$ 0.1} &  \nrm{96.1} {\cz $\pm$ 0.1} &  \nrm{96.3} {\cz $\pm$ 0.1} &  \nrm{96.3} {\cz $\pm$ 0.1} &  \nrm{96.1} {\cz $\pm$ 0.1} &  \nrm{96.7} {\cz $\pm$ 0.1} & \bst{96.9} {\cz $\pm$ 0.1} &  \bst{96.9} {\cz $\pm$ 0.1}  \\
        Election      &  \nrm{89.6} {\cz $\pm$ 0.6} &  \nrm{88.0} {\cz $\pm$ 0.6} &  \bst{90.8} {\cz $\pm$ 0.6} &  \nrm{90.5} {\cz $\pm$ 0.7} &  \nrm{87.3} {\cz $\pm$ 0.7} &  \nrm{85.4} {\cz $\pm$ 0.7} & \nrm{90.1} {\cz $\pm$ 0.8} &  \nrm{90.3} {\cz $\pm$ 0.9}  \\
        Sexual        &  \nrm{74.5} {\cz $\pm$ 1.4} &  \nrm{83.9} {\cz $\pm$ 1.2} &  \nrm{93.3} {\cz $\pm$ 0.8} &  \nrm{93.6} {\cz $\pm$ 0.9} &  \nrm{77.0} {\cz $\pm$ 1.7} &  \nrm{65.0} {\cz $\pm$ 1.1} & \nrm{97.1} {\cz $\pm$ 0.5} &  \bst{97.4} {\cz $\pm$ 0.4}  \\
    \bottomrule
    \end{tabular}
    }
    \label{tab:results_empirical_full_batch}
\end{table}

%
On the synthetic networks sampled from MRFs, the transductive GBPN outperforms the other methods by large margins (\Cref{tab:results_synthetic_full_batch}).
The performance gap is due to two reasons.
First, the underlying data assumption of our model exactly matches the data.
Second, GBPN is able to directly use training labels for inference.
On the empirical datasets, inductive and transductive GBPN outperform the baselines in five of seven graphs (\Cref{tab:results_synthetic_full_batch,tab:results_empirical_full_batch}).
GBPN is more accurate on most of the coauthorship and citation graphs, although the gains over the best baslines are modest.
The differences on the Election and Sexual datasets brings additional insights.
The node labels in both datasets are binary; however, 
election outcomes are homophilous, 
and the sexual interactions are heterophilous (most relationships used to construct the dataset are heterosexual).
GBPN is comparable to the best baseline on the Election network (GraphSAGE in this case), 
it outperforms the baselines by large margins on the Sexual network.
This is because GBPN can explicitly learn heterophily with the coupling-potential (\Cref{fig:coupling_potentials_f}), 
while the standard GNN models are designed for homophilous graphs~\cite{jia2020residual,zhu2020beyond}.
Finally, although the transductive GBPN consistently outperforms its inductive counterpart, the gap between them is small, as the predictive probabilities on the training nodes are already quite accurate.

\subsection{Identifiability, Interpretability and Convergence}
\begin{figure}[t]
    \phantomsubfigure{fig:coupling_potentials_a}
    \phantomsubfigure{fig:coupling_potentials_b}
    \phantomsubfigure{fig:coupling_potentials_c}
    \phantomsubfigure{fig:coupling_potentials_d}
    \phantomsubfigure{fig:coupling_potentials_e}
    \phantomsubfigure{fig:coupling_potentials_f}
    \phantomsubfigure{fig:coupling_potentials_g}
    \centering
    \includegraphics[width=1.00\linewidth]{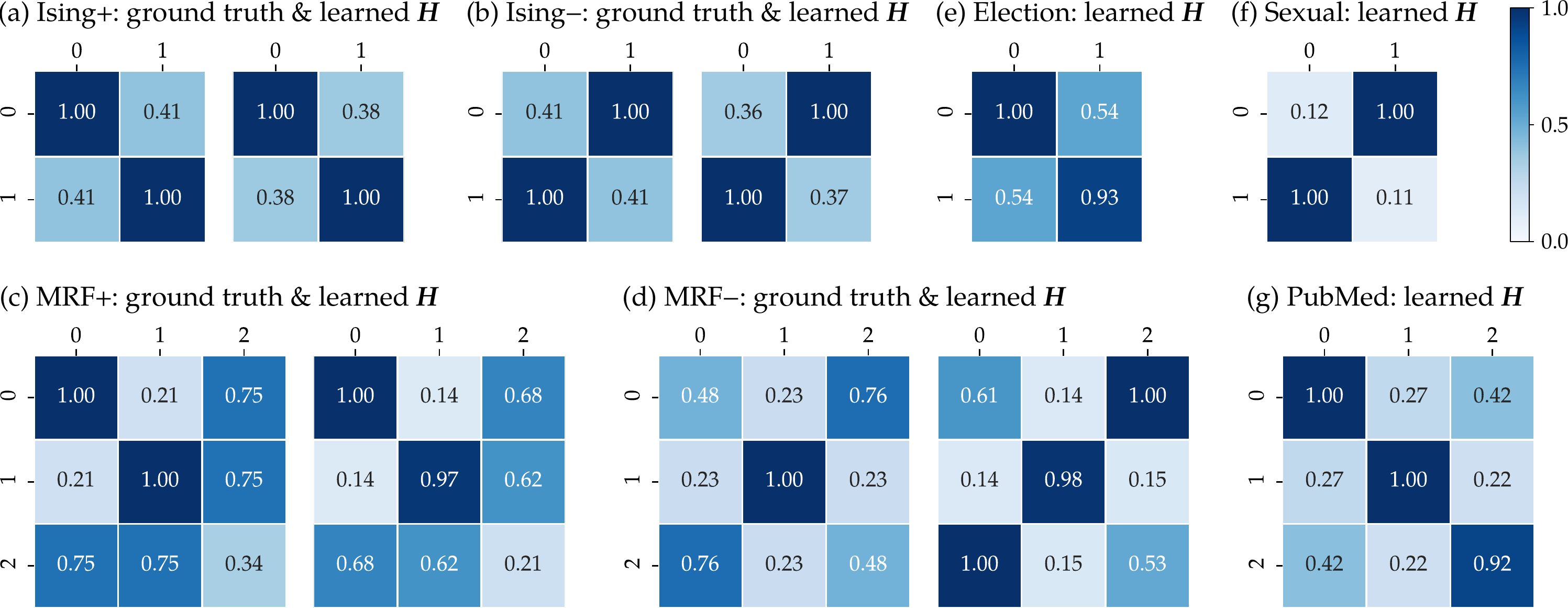}
    \caption{Visualization of the learned coupling matrices. 
    (a--d) Synthetic datasets, with ground truth coupling matrices on the left and learned coupling matrices on the right. 
    (e--g) Election, Sexual, and PubMed empirical datasets, where we recover homophilous, heterophilous,
    and mixed affinities.}
    \label{fig:coupling_potentials}
\end{figure}
A useful aspect of GBPN is that the learned coupling-potentials $\mH$ can be interpreted as between-class affinities.
We first confirm that GBPN identifies the true coupling-potentials used to generate the synthetic graphs.
%
%
Indeed, \Cref{fig:coupling_potentials_a,fig:coupling_potentials_b,fig:coupling_potentials_c,fig:coupling_potentials_d} shows that the learned $\mH$ is close to the parameters used to generate the data, although the exact numbers have minor differences.

Next, we examine the learned coupling matrices for some empirical datasets
(\Cref{fig:coupling_potentials_e,fig:coupling_potentials_f,fig:coupling_potentials_g}).
The learned coupling-potentials are homophilous on the Election network and strongly heterophilous in the Sexual network, which reflect the fact that the election outcome in neighboring counties tend to be the same and that most sexual interactions in the dataset are heterosexual.
On the PubMed citation network, node labels 0, 1, 2 correspond to medical publications on \texttt{experimental diabetes}, \texttt{adult-onset diabetes}, and \texttt{juvenile diabetes} respectively.
The learned coupling matrix shows that papers citing each other tend to be in the same field, as shown by the large diagonal values.
Beyond this, the smallest value is between \texttt{adult-onset diabetes} and \texttt{juvenile diabetes}.
This aligns with the fact that those two type of diseases are largely among different age groups.

%


%
\begin{figure}
    \phantomsubfigure{fig:pubmed_convergence_a}
    \phantomsubfigure{fig:pubmed_convergence_b}
    \centering
    \includegraphics[width=1.0\linewidth]{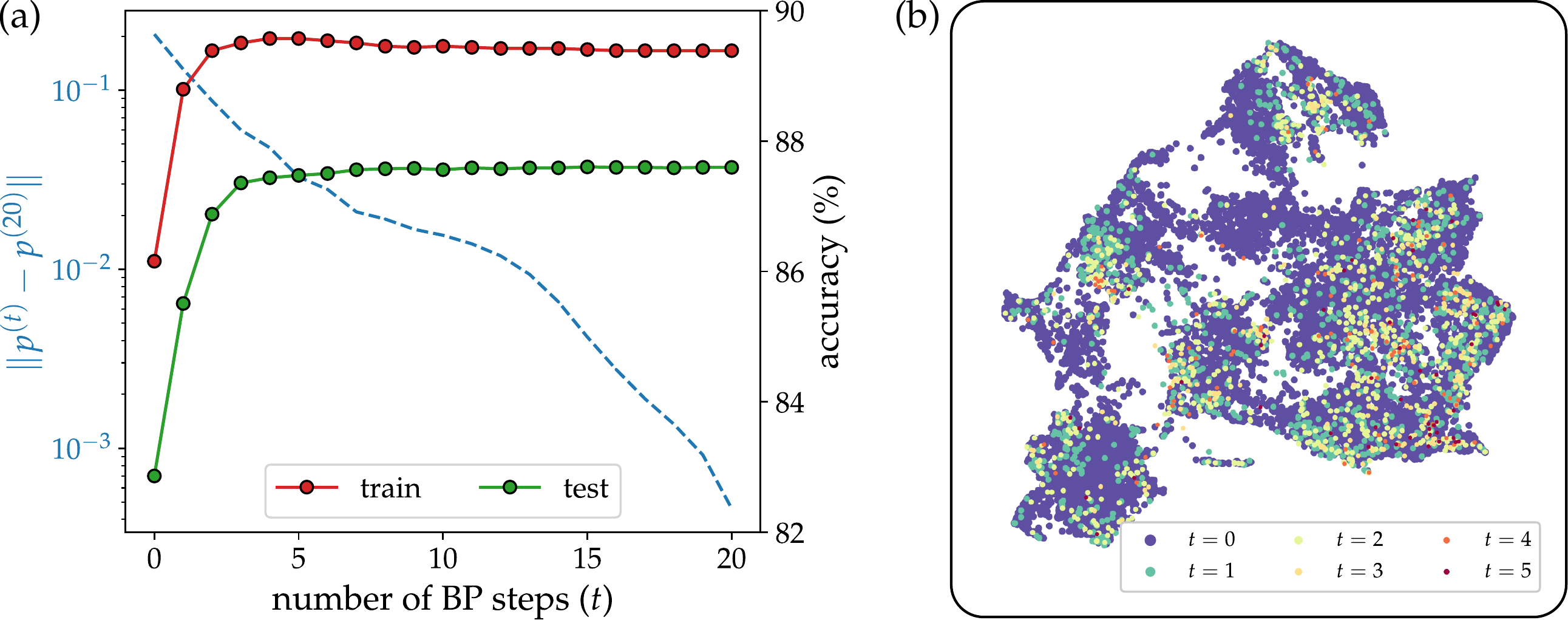}
    \caption{Convergence of GBPN on the PubMed dataset. 
    (a) Approximate residual error at step $k$ in the blue and train/test accuracies in red/green. 
    (b) Nodes correctly predicted with the GBPN algorithm, where the coordinates are given by a graph spectral embedding; 
    the color indicates the step $k$ after which the prediction is always correct.}
    \label{fig:pubmed_convergence}
\end{figure}

Another interesting component of the GBPN model is that we can examine intermediate predictions computed during belief propagation.
Although belief propagation is not guaranteed to converge on graphs with cycles, 
GBPN converged quickly on all real-world graphs, oftentimes in a handful of iterations. 
Here, we consider the PubMed dataset in detail, using the setup in \cref{subsec:results_full_batch}.
After training, we run belief propagation to compute the belief probability $\{p_{i}^{(t)}\}_{i \in V}$ for all nodes for up to $T=20$ steps.
To evaluate BP convergence, we measure 
the norm of the difference between the belief at step $t$ and the converged belief,
$\| p_{i}^{(t)} - p_{i}^{(\infty)}\| \approx \|p_{i}^{(t)} - p_{i}^{(20)} \|$,
along with the train and test accuracies (\Cref{fig:pubmed_convergence}).
%
The result shows that GBPN converges linearly with the number of belief-propagation steps, and the train and test accuracies change little after a handful of iterations.

To help understand BP convergence and the algorithm in general, we also visualize how each BP step refines the final prediction (\Cref{fig:pubmed_convergence}).
We plot the nodes that are correctly predicted by GBPN at step $5$, 
where the coordinate of each node is computed by compressing its spectral embeddings~\cite{belkin2001laplacian} and features together with UMAP~\cite{mcinnes2018umap}, 
and the color indicates the step after which the prediction is always correct.
Nodes corrected at later steps (e.g., $t=5$ in dark red) are typically surrounded by others examples that are already corrected earlier.
This behavior reflects the mechanism of the GBPN model: start with easy data points and iteratively correct harder examples with neighbor information.

\subsection{Neighborhood Sub-sampling and Performance on Larger Datasets \label{subsec:subsampling}}
\begin{figure}[t]
    \phantomsubfigure{fig:misalignment_a}
    \phantomsubfigure{fig:misalignment_b}
    \phantomsubfigure{fig:misalignment_c}
    \centering
    \includegraphics[width=1.0\linewidth]{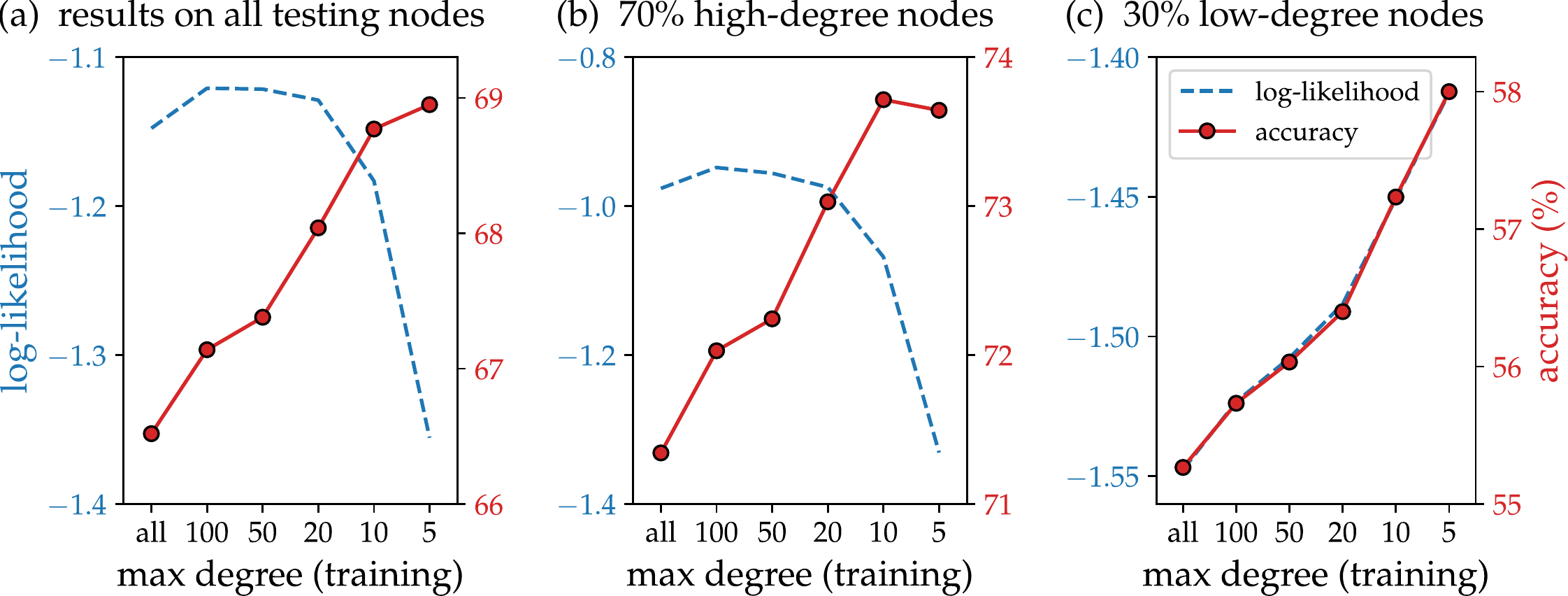}
    \caption{Visualization of the misalignment between the (unweighted) negative log-likelihood loss function and prediction accuracy on a graphs with high degree heterogeneity. (a) when training GBPN with degree regularization, the predictive log-likelihood on the testing nodes decreases while accuracy increases. (b) this misalignment is caused by 70\% of nodes with the highest degrees. (c) on the rest 30\% of low-degree nodes, the log-likelihood aligns perfectly with accuracy.}
    \label{fig:misalignment}
\end{figure}
In \Cref{subsec:practical_considerations}, we identified the misalignment between the unweighted NLL loss function and the classification accuracy on graphs with high degree heterogeneity, which prompts us to down-weight the high degree nodes.
Here, we consider neighborhood sub-sampling as an alternative solution to the misalignment problem, which also reduces computation and enables efficient mini-batch training.

To demonstrate how degree heterogeneity causes the misalignment and consequently hurts accuracy, we manually reduced the degree heterogeneity by regularizing the high degree nodes during training.
In particular, we train a 1-hop GBPN on the OGBN-arXiv dataset with the unweighted NLL loss, 
uniformly sub-sampling a maximum number of neighbors during the aggregation step in \cref{eq:gbpn_aggregation}
(or all neighbors if the degree is below the threshold).
At inference, we run GBPN with all neighbors to predict the testing labels.
\Cref{fig:misalignment_a} shows the average log-likelihood and accuracy on the testing nodes as a function of the maximum number of neighbors, and reducing this number improves the testing accuracy.
At the same time, the log-likelihood of the testing node decreases, showing the misalignment.
We further split the testing nodes into the $30\%$ lowest degree nodes and the rest $70\%$, 
and find that the log-likelihood and accuracy aligns on the low-degree nodes (\Cref{fig:misalignment_b})
but not on the high-degree nodes (\Cref{fig:misalignment_c}).
This justifies our heuristic of down-weighting high degree nodes.

\begin{figure}[t]
    \centering
    \includegraphics[width=1.0\linewidth]{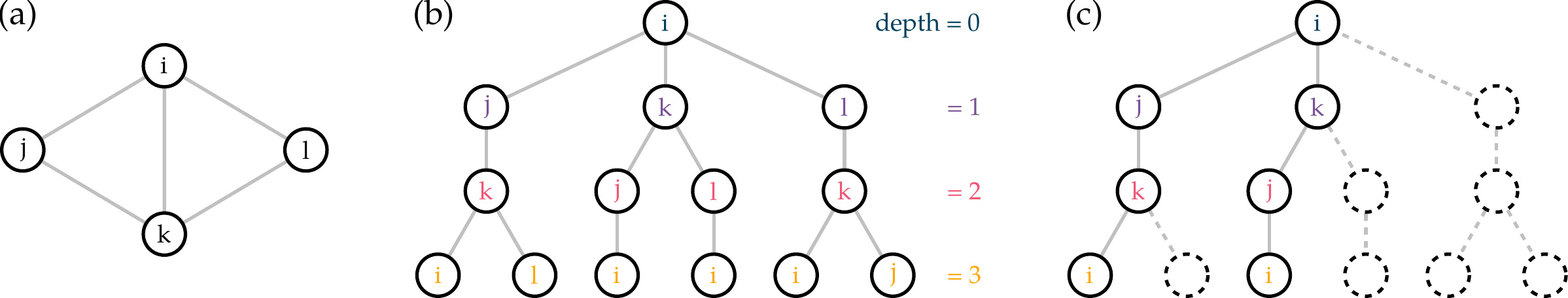}
    \caption{Demonstration of the sub-sampling algorithm on a simple graph. (a) the original graph. (b) the unrolled tree from the root node $u$. (c) the sub-sampled tree when the maximum degree is $2$.}
    \label{fig:subsampling}
\end{figure}
\begin{table}[t]
    \caption{Node classification performances on large-scale empirical datasets with mini-batch training.}
    \centering
    \resizebox{0.86\textwidth}{!}{
    \begin{tabular}{r @{\qquad}  llll @{\qquad} ll}
    \toprule
    dataset      &  MLP                        &  ClusterGCN                 &  SAGE                       &  GAT                        &  GBPN-I                     &  GBPN                        \\
    \midrule                                                                 
    Elliptic     &  \nrm{88.4} {\cz $\pm$ 0.3} &  \nrm{81.4} {\cz $\pm$ 0.6} &  \nrm{88.5} {\cz $\pm$ 0.5} &  \nrm{88.3} {\cz $\pm$ 0.3} &  \nrm{90.0} {\cz $\pm$ 0.7} &  \bst{90.1} {\cz $\pm$ 0.7}  \\
    Payments     &   \nrm{6.9} {\cz $\pm$ 0.5} &  \nrm{40.7} {\cz $\pm$ 1.5} &  \nrm{26.6} {\cz $\pm$ 1.2} &  \nrm{19.0} {\cz $\pm$ 1.2} &  \nrm{47.9} {\cz $\pm$ 1.0} &  \bst{53.2} {\cz $\pm$ 1.3}  \\
    arXiv        &  \nrm{54.9} {\cz $\pm$ 0.3} &  \nrm{66.4} {\cz $\pm$ 0.5} &  \nrm{70.2} {\cz $\pm$ 0.1} &  \nrm{70.1} {\cz $\pm$ 0.4} &  \nrm{70.7} {\cz $\pm$ 0.3} &  \bst{71.8} {\cz $\pm$ 0.3}  \\
    Products     &  \nrm{61.3} {\cz $\pm$ 0.1} &  \nrm{77.5} {\cz $\pm$ 0.7} &  \nrm{78.6} {\cz $\pm$ 0.2} &  \nrm{74.8} {\cz $\pm$ 2.2} &  \bst{81.8} {\cz $\pm$ 0.3} &  \bst{81.8} {\cz $\pm$ 0.2}  \\
    \midrule
    \end{tabular}
    }
    \label{tab:results_empirical_mini_batch}
\end{table}
For large-scale graphs where mini-batch training is necessary, neighborhood sub-sampling improves the classification accuracy
and also reduces computational costs.
In particular, computing the $T$-step belief of a node $i$ requires unrolling a computation tree~\cite{ihler2005loopy} that includes all $T$-hop neighbors of $i$, which can quickly grow to the entire graph.
With neighborhood sub-sampling, we randomly select a maximum of $d$ neighbors per node when unrolling the tree, 
and the sub-sampled tree has $\mathcal{O}(d^{T})$ nodes (\Cref{fig:subsampling}).
This is similar to sub-sampling in GNNs to address scalability~\cite{hamilton2017inductive}, 
although our sample is a tree and also explicitly down-weights high-degree nodes.
%
%

We test the performance of GBPN with mini-batch training on four large-scale real-world datasets.
The experimental setup is the same as in \Cref{subsec:results_full_batch},
except we only train for $100$ epochs, as the model parameters are updated multiple times per epoch.
We repeat each experiment only $10$ times because accuracy has smaller variance.
During GBPN training, we use $T = 2$ BP steps and a maximum of $5$ sampled neighbors.
Training on the largest dataset (Products) takes about one minute per epoch and less than two hours overall on a Telsa V100 GPU.
For a fair comparison, we use a similar neighborhood expansion method for mini-batch training of GraphSAGE and GAT, 
with the branching factor set to $5$.
We replace the GCN baseline with ClusterGCN~\cite{chiang2019clustergcn}, where we set the average size per cluster to be $256$ and the number of clusters per mini-batch to be $3$; other hyperparameters are kept the same.
We do not include GMNN and DeeperGNN as baselines since neither of the reference repositories implemented mini-batch training.
The labels in the Elliptic and Payments datasets are highly imbalanced as most transactions are licit, 
so we measure the prediction performance with the F1 score ($\times 100$) on the illicit class.
Performance on the arXiv and Products datasets is measured by the classification accuracy.
We find that GBPN-I outperforms the baselines on all four datasets (\Cref{tab:results_empirical_mini_batch}),
and GBPN provides additional gains on the Payments and arXiv datasets.

In addition to uniform sampling, we also experimented with importance sampling, which has proven useful for training certain graph neural networks~\cite{zhang2021biased,liu2020bandit}.
The results are summarized in \Cref{sec:importance_sampling}.


\section{Conclusions and Future Work \label{sec:conclusions}}
We investigated belief propagation as a model component for node classification, developing a model that combines the advantages of graph neural networks and collective classification.
Our model is easy-to-train, accurate, and scalable while maintaining interpretability and having a natural way to incorporate training labels for inference.
Results on real-world datasets justify our claims.
There are several fruitful directions for future research.
For instance, since BP often converges to a fix point, implicit differentiation could be used to reduce memory consumption~\cite{bai2019deep}.
One could also extend edge potentials to motif potentials to incorporate higher order interactions.
%
Finally, although our algorithms are generic, improvements in node classification could lead to more personalized ad targeting or reasoning about sensitive information, which could have negative societal impact.

\xhdr{Acknowledgements}
This research was supported by NSF award DMS-1830274, ARO award W911NF-19-1-0057, ARO MURI, and JPMorgan Chase \& Co.

\xhdr{Disclaimer}
This paper was prepared for information purposes by the AI Research
Group of JPMorgan Chase \& Co and its affiliates (``J.P. Morgan''),
and is not a product of the Research Department of J.P. Morgan. J.P.
Morgan makes no explicit or implied representation and warranty and
accepts no liability, for the completeness, accuracy or reliability of
information, or the legal, compliance, financial, tax or accounting
effects of matters contained herein. This document is not intended
as investment research or investment advice, or a recommendation,
offer or solicitation for the purchase or sale of any security, financial
instrument, financial product or service, or to be used in any way for
evaluating the merits of participating in any transaction.

\bibliographystyle{unsrt}
\bibliography{main}



\appendix
\section{Supplementary Materials for Reproducing the Main Results \label{sec:reproducibility}}
Here we provide some implementation details of our methods to help readers reproduce and further understand the algorithms and experiments in this paper.

\subsection{Additional Details on Implementations \label{subsec:additional_implementation}}
\begin{algorithm}[ht]
\SetNoFillComment
\SetKwInOut{Input}{Input}
\SetKwInOut{Output}{Output}
\Input{\ graph topology $G(V,E)$; node features $\{\vx_{i}\}_{i \in V}$; number of BP iterations $T$; \\ \ self-potential mapping $g_{\theta}$; coupling matrix $\mH$}
\BlankLine
\Output{\ prediction for each node label $\{\hat{y}_{i}\}_{i \in V}$}
\BlankLine
\tcc{step 0: initialize messages}
\For{$(i,j) \in E$}{
    \For{$y_{j} = 0 \ldots (c-1)$}{
        $\log m_{ij}^{(0)}(y_{j}) \gets -\log c$
    } 
    \BlankLine
    \For{$y_{i} = 0 \ldots (c-1)$}{
        $\log m_{ji}^{(0)}(y_{i}) \gets -\log c$
    }
}
\BlankLine
\tcc{step 1: initialize beliefs}
\For{$i \in V$}{
    \For{$y_{i} = 0 \ldots (c-1)$}{
        $\log p_{i}^{(0)}(y_{i}) \gets \log g_{\theta}(y_{i}; \vx_{i})$
    }
}
\BlankLine
\tcc{step 2: belief propagation iterations}
\For{$t = 1 \ldots T$}{
    \For{$(i,j) \in E$}{
        \For{$y_{j} = 0 \ldots (c-1)$}{
            $\log m_{ij}^{(t)}(y_{j}) \leftlsquigarrow \texttt{LSE}_{y_{i}}\left[\log H_{y_{i},y_{j}} + \log p_{i}^{(t-1)}(y_{i}) - \log m_{ji}^{(t-1)}(y_{i})\right]$
        } 
        \BlankLine
        \For{$y_{i} = 0 \ldots (c-1)$}{
            $\log m_{ji}^{(t)}(y_{i}) \leftlsquigarrow \texttt{LSE}_{y_{j}}\left[\log H_{y_{j},y_{i}} + \log p_{j}^{(t-1)}(y_{j}) - \log m_{ij}^{(t-1)}(y_{j})\right]$
        }
    }
    \BlankLine
    \For{$i \in V$}{
        \For{$y_{j} = 0 \ldots (c-1)$}{
            $\log p_{i}^{(t)}(y_{i}) \leftlsquigarrow \log p_{i}^{(0)}(y_{i}) + \sum_{j \in N(i)} \log m_{ji}^{(t)}(y_{i})$
        }
    }
}
\BlankLine
\tcc{step 3: final predictions}
\For{$i \in V$}{
    $\hat{y}_{i} = \arg\max_{y_{i}} p_{i}^{(T)}(y_{i})$
}
\caption{The inference (i.e. forward propagation) algorithm for inductive GBPN.}
\label{alg:gbpni}
\end{algorithm}
Here, we summarize the inductive GBPN variant in \cref{alg:gbpni}, where $\leftlsquigarrow$ denotes ``assign after normalization.''
In contrast, the transductive GBPN initializes the belief $\log p_{i}^{(0)}(y_{i})$ on each training node $i \in L$ to be $\log p_{i}^{(0)}(y_{i}) = 0$ if $y_{i}$ is the ground-truth class and $\log p_{i}^{(0)}(y_{i}) = -\infty$ otherwise.
\subsection{Additional Details on Experimental Setup\label{subsec:additional_setup}}
\xhdr{Architectures}
The first step of GBPN is to map the features on each node into a $c$-dimensional initial belief (or self-potential) vector using a MLP.
For this MLP, we use a 2-hidden-layer feedforward network with 256 hidden units and ReLU activation function.
During training, on each dataset with a small feature dimension relative to the number of training nodes (Ising, MRF, County, Sexual, Elliptic, Payment, arXiv, Products), we set the dropout probability to $0.1$; 
on each dataset with a large feature dimension relative to the number of training nodes (Cora, CiteSeer), we set the dropout probability to $0.6$;
on the rest of the datasets (PubMed, CS, Physics), we set the dropout probability to $0.3$.
The same activation function and dropout probabilities are used for baseline methods.

\xhdr{Optimization}
For all methods except for GMNN and DeeperGNN, the parameters are optimized using a AdamW optimizer with with $\beta_{1} = 0.9$, $\beta_{2} = 0.999$, learning rate $1.0 \times 10^{-3}$, and decay rate $2.5 \times 10^{-4}$.
All experiments are performed on a server with a Xeon 6254 CPU and a Telsa V100 GPU.

\subsection{Additional Details on Datasets \label{subsec:additional_datasets}}
\xhdr{Synthetic MRF datasets}
We sample random configurations from MRFs defined on a $2$-dimensional grid graph.
A MRF configuration assigns a label to every node in the graph.
The MRF models we use consist of two types of potentials: 
(i) the self-potential on each node (ii) the coupling-potential between each pairs of neighboring nodes.
While the coupling-potential for each dataset is shown in \cref{fig:coupling_potentials}, here we describe how we set the self-potentials on each node.
Let $r_{i1}$ and $r_{i2}$ denote the first and second grid coordinates of node $i$ normalized between $-1.0$ and $+1.0$, the self-potential $h_{i}(y_{i})$ on each node $i$ is defined as a function of the coordinates.
\begin{itemize}[leftmargin=0.2in]
    \vspace{-0.07in}
    \item \textbf{Ising$\mathbf{+}$:} $h_{i}(0) = \exp\left(-0.35 \cdot r_{i1} \cdot r_{i2}\right)$, $h_{i}(1) = \exp\left(+0.35 \cdot r_{i1} \cdot r_{i2}\right)$
    \item \textbf{Ising$\mathbf{-}$:} $h_{i}(0) = \exp\left(-0.35 \cdot r_{i1} \cdot r_{i2}\right)$, $h_{i}(1) = \exp\left(+0.35 \cdot r_{i1} \cdot r_{i2}\right)$
    \item \textbf{MRF$\mathbf{+}$:} $h_{i}(y_{i}) = \texttt{sigmoid}(0.2 \cdot s_{i}(y_{i}))$; \enskip $s_{i}(0) = 0$, $s_{i}(1) = r_{i1}^{2} + r_{i2}^{2} - 0.65$, $s_{i}(2) = -s_{i}(1)$
    \item \textbf{MRF$\mathbf{-}$:} $h_{i}(y_{i}) = \texttt{sigmoid}(0.6 \cdot s_{i}(y_{i}))$; \enskip $s_{i}(0) = 0$, $s_{i}(1) = r_{i1}^{2} + r_{i2}^{2} - 0.00$, $s_{i}(2) = -s_{i}(1)$
    \vspace{-0.07in}
\end{itemize}
We use the Metropolis algorithm for simulating the Ising models and Gibbs sampling~\cite{mezard2009information} for simulating the $3$-class MRF models.
The synthetic datasets are released together with our GBPN implementation.

\section{Importance Sampling \label{sec:importance_sampling}}
In \cref{subsec:subsampling}, we observed that regularizing high degree nodes during training --- by aggregating messages across a small and uniformly sampled neighborhood $N'(i)$ rather than the entire neighborhood $N(i)$ ---
leads to a boost in both accuracy and computational efficiency.
However, one potential concern is that estimation with uniformly sampled neighbors has high variance.
To this end, we explore importance sampling to reduce the variance.
Surprisingly, we find the optimal sampling distribution that minimizes the variance is very close to uniform sampling on benchmark datasets.

\subsection{Theoretical Optimal Distribution for Importance Sampling \label{subsec:opt_dist}}
We first formulate the neighbor sub-sampling procedure according to an arbitrary sampling distribution $p \in [0,1]^{|N(i)|}$ by considering a sequence of $d \leq |N(i)|$ independent and identically distributed (i.i.d.) draws of neighbor indices, $N'(i) = (j_1, \ldots, j_{d})$, where each $j$ is sampled according to $p$. 
For each of the samples $j \in N'(i)$, let 
\begin{align}
X_j(y_i) = \log m_{ji}^{(k)}(y_i)
\end{align}
be the random variable corresponding to the message from $j \in N'(i)$ for class $y_i$. Further denote the sum of the incoming messages over the sampled neighbors with respect to class $y_i$ as $X(y_i)$, i.e.,
\begin{align}
X(y_i) = \sum_{j \in N'(i)} X_j(y_i).
\end{align}

Denote the joint distribution of all $d$ samples as $p^{d}$, we can compute the expectation of $X(y_{i})$,
\begin{align*}
\E_{N'(i) \sim p^d}[X(y_i)] &= |N'(i)| \E_{j \sim p}[X_j(y_i)] &(\text{linearity of expectation \& i.i.d. sampling})\\
&= |N'(i)| \sum_{j \in N(i)} \left(\log m_{ji}^{(k)}(y_i)\right) p_j &(\text{definition of $\E_{j \sim p}[X_j(y_i)]$}) \\
&= \frac{|N'(i)|}{|N(i)|} \sum_{j \in N(i)} \log m_{ji}^{(k)}(y_i). &(\text{$p_j = \nicefrac{1}{|N(i)|}$ for uniform sampling})
\end{align*}
Hence, $X(y_i)$ is an unbiased estimator for the quantity 
\begin{align}
\label{eq:target}
\Omega(y_{i}) = \frac{|N'(i)|}{|N(i)|} \sum_{j \in N(i)} \log m_{ji}^{(k)}(y_i),
\end{align}
which represents the scaled aggregate messages from all neighbors $N(i)$. 
However, the estimation given by $X(y_{i})$ may have high variance.
Thus, we use an unbiased and low-variance estimator defined by importance sampling.
In particular, we consider the following estimator:
\begin{align}
    Z(y_{i}) = \sum_{j \in N'(i)} Z_j(y_i) = \sum_{j \in N'(i)} \frac{X_j(y_i)}{|N(i)| \cdot p_j},
\end{align}
which simply re-weights $X_{j}(y_{i})$.
It is easy to show $Z(y_{i})$ is an unbiased estimator for $\Omega(y_{i})$:
\begin{align}
\E_{N'(i) \sim p^d} [ Z(y_i)] = |N'(i)| \E_{j \sim p} [ Z_j(y_i)] &= \frac{|N'(i)|}{|N(i)|} \E_{j \sim p} [ X_j / p_i] \\
&= \frac{|N'(i)|}{|N(i)|} \sum_{j \in N(i)} \log m_{ji}^{(k)}(y_i).
\end{align}

Now, our goal is to find a probability distribution $p^* \in [0,1]^{|N(i)|}$ so that $Z(y_{i})$ has smallest possible variance, i.e.,
\begin{align}
    p^* &= \argmin_{p} \mathrm{Var}_{p^d} (Z(y_i)),
\end{align}
where $\mathrm{Var}_{p^d} (Z(y_i))$ is the variance of $Z(y_i)$ under the sampling distribution $p^d$. 
The variance further simplifies to
\begin{align}
    \mathrm{Var}_{p^d} (Z(y_i)) &= |N'(i)| \mathrm{Var}_{p} (Z_j(y_i)) \nonumber &(\text{i.i.d. sampling}) \nonumber \\
    &= \frac{|N'(i)|}{|N(i)|^2} \mathrm{Var}_{p} (X_j(y_i) / p_j) &(\text{definition of $Z_j(y_i)$}) \nonumber\\
    &= \frac{|N'(i)|}{|N(i)|^2} \left(\E_{p}[(X_j(y_i)/p_j)^2] - \E_{p}[X_j(y_i)/p_j]^2 \right). \nonumber
    \label{eq:var}
\end{align}
By unbiasedness, we have that $(\E[X_j(y_i)/p_j])^2 = \left(\sum_{j \in N(i)} \log m_{ji}^{(k)}(y_i) \right)^2$, which implies that the distribution $p$ that minimizes the variance is the one which minimizes the second moment $\E_{j \sim p}[(X_j(y_i)/p_j)^2]$. This evaluates to
\begin{align}
\E_{j \sim p}[(X_j(y_i)/p_j)^2] = \sum_{j \in N(i)} \left(\log m_{ji}^{(k)}(y_i)\right)^2/p_j.
\end{align}
The expression above is convex in $p$, so using the method of Lagrange multipliers,\footnote{More generally, the distribution $p_i = |x_i|/\sum_j |x_j|$ minimizes the sum $\sum_{i} x_i^2/p_i$.} we find that the minimizer is given by
\begin{align}
p^*_j \propto |\log m_{ji}^{(k)}(y_i)| \quad \forall{j \in |N(i)|}.
\end{align}

Since we want a distribution that minimizes the variance over all potential classes $y_i \in \{0, \ldots, c-1\}$, we further consider minimizing the sum of variances over all classes
\begin{align}
p^* = \argmin_{p} \sum_{y_i} \mathrm{Var}_{p^d} (Z(y_i)).
\end{align}
When the number of samples is clear from context, we will let
\begin{align}
\var(p) = \sum_{y_i} \mathrm{Var}_{p^d} (Z(y_i))
\end{align}
denote the sum of variances over all classes under distribution $p$. An analogous derivation leads to the optimal distribution having the form
\begin{equation}
\label{eq:opt-p}
p_j^* \propto \sqrt{ \sum_{y_i} (\log m_{ji}^{(k)}(y_i) )^2}.
\end{equation}






\subsection{Online Learning}
We would ideally like to sample the neighbors according to the optimal sampling distribution as defined above, but the optimal distribution requires knowledge of all of the neighbors' messages. Since we do not know the messages prior to sampling, we are faced with the dilemma where we cannot sample optimally prior to seeing the messages, but cannot see the messages until we sample. We can address via the well-studied exploration-and-exploitation trade-off in online learning with partial information, which includes the Multi-armed Bandit setting~\cite{orabona2019modern,uchiya2010algorithms}.



We formulate the variance minimization problem as an online learning problem where the goal is to minimize the regret (sum of variances) with respect to the best sampling distribution in hindsight. In particular, we consider the linearized, partial-information setting where the losses at each epoch $\ell_{\tau}$ correspond to the gradient of the variance, i.e., for each $j \in N'(i)$, 
\begin{align}
\label{eq:loss-def}
\ell_{\tau,j} = - \frac{\sqrt{\sum_{y_i} (\log m_{ji}^{(k)}(y_i))^2}}{p_{\tau,j}^2},
\end{align}
and $\ell_{\tau,j} = 0$ for all $j \notin N'(i)$,
where $p_{\tau,j}$ is the probability of sampling neighbor $j$ at epoch $\tau$.
Note that we are in the partial information setting since we do not see the messages of all neighbors, but only those of sampled neighbors, so we will need to reweigh the losses $\ell_{\tau,j}$ by an additional $p_{\tau,j}$ term in the denominator so that we can approximate the full gradient in expectation (see~\cite[Algorithm 10.2]{orabona2019modern} for details). 
Our overarching goal is to bound the expected regret with respect to the best distribution in hindsight,
\begin{align}
\mathrm{Regret}(\{p_{\tau}\}) = \sum_{\tau} \mathrm{Var}(p_\tau) - \min_{p \in \Delta} \sum_{\tau} \mathrm{Var}(p).
\end{align}
where $p_\tau \in \Delta$ is the sampling distribution at epoch $\tau$ and $\Delta$ is the probability simplex. 
By convexity of the variance with respect to the probabilities $p$, the regret defined above is upper bounded by the following linearized regret, i.e.,
\begin{align}
\mathrm{Regret}(\{p_{\tau}\}) \leq \sum_{\tau} \dotp{p_\tau}{\ell_\tau} - \min_{p \in \Delta} \sum_{\tau} \dotp{p}{\ell_\tau},
\end{align}
Hence, minimizing this upper bound would guarantee that the sequence of sampling distributions we used in sampling the neighbors of a node $i$ is competitive with the best sampling distribution in hindsight. It can be shown~\cite[Theorem 10.2]{orabona2019modern} that for an appropriate choice of the learning rate, the regret of the \textsc{Exp3} algorithm with respect to sampling the neighbors of node $i$ is bounded by
$\mathcal{ O}\left(\sqrt{\log (|N(i)|) \, \sum_{\tau} \E_{j \sim p}[\ell_{\tau,j}^2}] \right)$.

\subsection{Empirical Evaluations \& Discussion}

\begin{figure*}[htb!]
  \centering
  \begin{minipage}[t]{0.47\textwidth}
    \centering
    \includegraphics[width=1.00\textwidth]{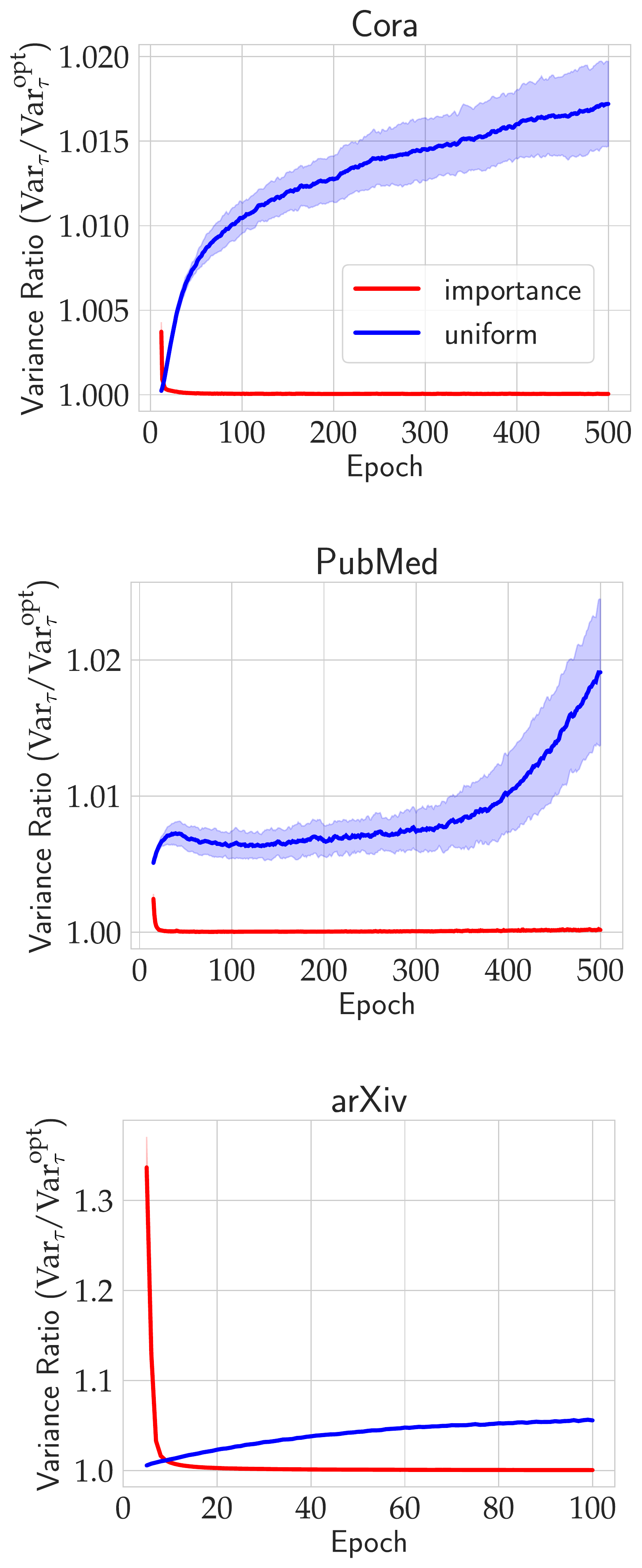} 
    \subcaption{Variance Relative to Optimal ($\var_\tau^\mathrm{opt}$) \label{fig:variance_a}}
  \end{minipage}%
  \hfill
  \begin{minipage}[t]{0.47\textwidth}
    \centering
    \includegraphics[width=1.00\textwidth]{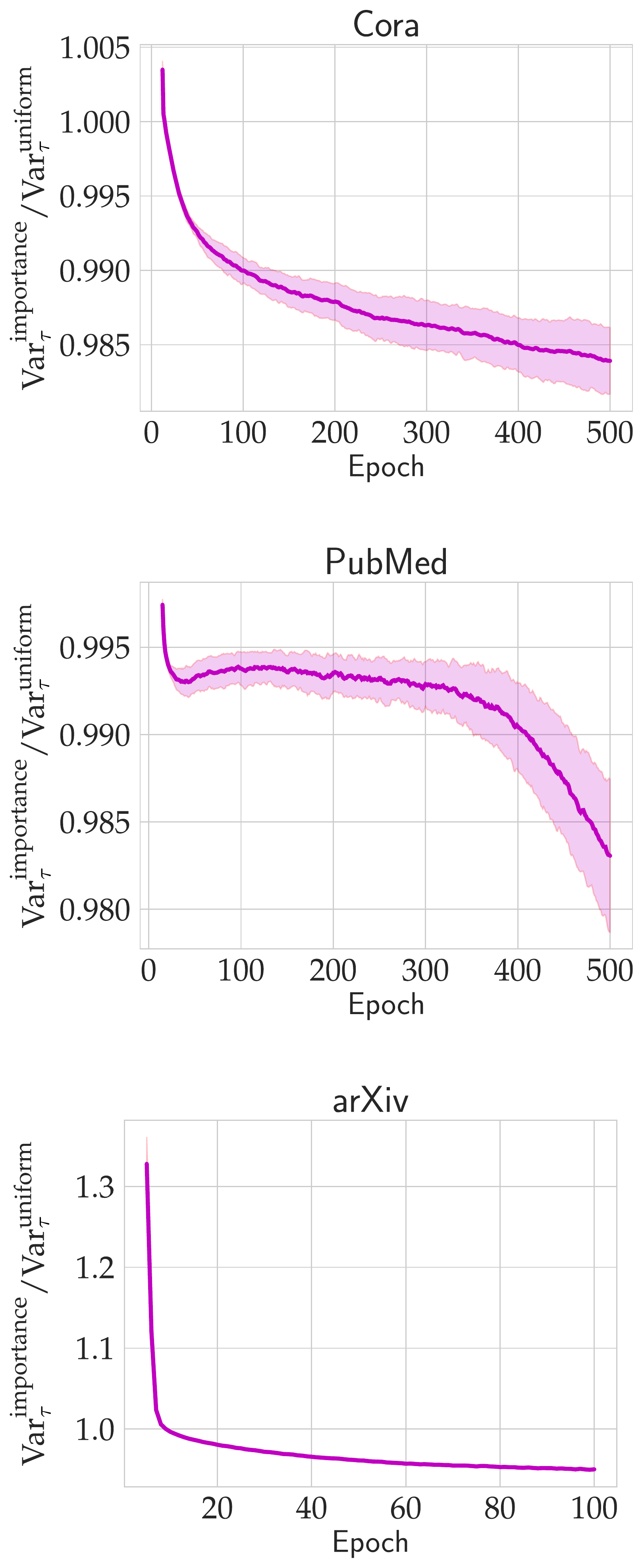}
    \subcaption{Variance of \textsc{Importance} relative to \textsc{Uniform} \label{fig:variance_b}}
  \end{minipage}%
\caption{Evaluations of the relative variance under our sampling distribution and uniform sampling on varying data sets. The ratio of the variances are averaged over all nodes per trial and shaded regions correspond to the values within one standard deviation from the mean.}
	\label{fig:variance}
\end{figure*}

We implemented the adaptive version of \textsc{Exp3} with time-varying learning rates~\cite[Chapter 7.6]{orabona2019modern}; other variants of \textsc{Exp3} provide similar regret guarantees~\cite{zhang2021biased,liu2020bandit}. Across evaluations on the data sets described in \Cref{tab:data_statistics}, we observed very similar performance between our regret-based, importance sampling approach and uniform sampling. 
Even for cases where importance sampling with \textsc{Exp3} improved the accuracy, these improvements were not significant in our evaluations, i.e., they were well-within one standard deviation of uniform sampling's performance. 

To dig deeper into our (unexpected) findings, we questioned whether \textsc{Exp3} (and importance sampling) was providing a variance reduction at all. To this end, we plotted the variance of the distribution generated by \textsc{Exp3} and compared it to the variance under the optimal sampling distribution $p^*_\tau$ in hindsight (computed as in \cref{eq:opt-p}) after each training epoch $\tau$. We performed the calculation for uniform sampling. Note that unlike in the regret definition above, $p^*_\tau$ is the optimal distribution \emph{with respect to epoch $\tau$}, meaning that it has the lowest variance for epoch $\tau$ possible, $\var_\tau^\mathrm{opt}$. We also directly compared the variance of our importance sampling distribution with that of uniform sampling, to determine whether the similarity in performance was a result of similar estimator variance.

\Cref{fig:variance} depicts the results of our evaluations on the Cora, PubMed, and arXiv datasets. In \Cref{fig:variance_a}, we plot the ratio for the variances under \textsc{Importance} and \textsc{Uniform} distributions relative to the variance $\var_\tau^\mathrm{opt}$ under the optimal distribution $p_\tau^*$. We observe that after a brief burn-in period, the variance ratio of \textsc{Importance} is very close to 1, implying that \textsc{Exp3} in fact generates sampling distributions that are essentially optimal ($p_\tau^*$) for each time step. In fact, we see that as time progresses, our online learning algorithm is able to learn increasingly better (closer to optimal) distributions. On the other hand, uniform sampling's performance relative to the optimal sampling distribution tends to degrade over the training process. Similarly, in \cref{fig:variance_b}, where we directly compute the ratio of variances between \textsc{Importance} and \textsc{Uniform}, i.e.,
\begin{align}
\rho_\tau = \frac{\var_\tau^\mathrm{importance}}{\var_\tau^\mathrm{uniform}},
\end{align}
we see \textsc{Importance} increasingly outperforms \textsc{Uniform} $(\rho_\tau < 1)$ after the initial 10-20 epochs.

Despite the qualitatively encouraging results, the scales of the y-axis in \cref{fig:variance} help us understand why \textsc{Importance} does not ultimately lead to better performance in terms of test accuracy. In particular, \Cref{fig:variance} shows that we achieve at most only $5\%$ reduction in variance relative to uniform sampling throughout virtually the entire training process. These empirical evaluations suggest that uniform sampling is very close to the per-time-step optimal distribution $p_\tau^*$ and provide an explanation for the lack of statistically significant improvement in test accuracy we observed with importance sampling.  

We envision that practitioners can bridge online learning with our framework and use, e.g., \textsc{Exp3} as we described it here, on scenarios where uniform sampling of neighbors may not be close to optimal. By bridging tools with theoretical guarantees from online learning with our work, we can in effect bound the worst-case performance of our sub-sampling procedure against even adversarial inputs. In future work, we plan to investigate scenarios where importance sampling can in fact yield significant improvements when it comes to both variance reduction and increased test accuracy.

\end{document}